\documentclass[review]{elsarticle}

\usepackage{lineno,hyperref}
\modulolinenumbers[5]

\usepackage{multirow}

\journal{arXiv}

\usepackage{caption}
\usepackage{subcaption}
\usepackage{amsmath}
\usepackage[ruled,vlined]{algorithm2e}
\include{pythonlisting}

\usepackage{algcompatible}
\usepackage{color}

\algnewcommand\algorithmicdata{\textbf{Data:}}
\algnewcommand\DATA{\item[\algorithmicdata]}

\algnewcommand\algorithmicparameters{\textbf{Parameters:}}
\algnewcommand\PARAMETERS{\item[\algorithmicparameters]}

\algnewcommand\algorithmicbegin{\textbf{begin}}
\algnewcommand\BEGIN{\item[\algorithmicbegin]}

\algnewcommand\algorithmicEND{\textbf{end}}
\algnewcommand\END{\item[\algorithmicEND]}

\usepackage{amsthm}

\theoremstyle{plain}
\newtheorem{thm}{Theorem}[section] 

\theoremstyle{definition}
\newtheorem{defn}[thm]{Definition} 

\usepackage{xcolor}

\newcommand{\new}[1]{{\color{black}#1}}

\begin{document}

\begin{frontmatter}

\title{\textit{How fair can we go in machine learning?}\\
Assessing the boundaries of fairness in decision trees}

\author{Ana Valdivia}
\address{Trilateral Research, London SW1X 7QA, United Kingdom}
\fntext[myfootnote]{ana.valdivia@trilateralresearch.com (corresponding author)}

\author{Javier S\'anchez-Monedero}
\address{Data Justice Lab, Cardiff University, Cardiff CF10 1FS, United Kingdom}
\fntext[myfootnote2]{sanchez-monederoj@cardiff.ac.uk}

\author{Jorge Casillas}
\address{Department of Computer Science and Artificial Intelligence, University of Granada, 18071 Granada, Spain}
\fntext[myfootnote3]{casillas@decsai.ugr.es}






\begin{abstract}
Fair machine learning works have been focusing on the development of equitable algorithms that address
discrimination of certain groups. Yet, many of these fairness-aware approaches aim to obtain a unique
solution to the problem, which leads to a poor understanding of the statistical limits of bias mitigation interventions.

We present the first methodology that allows to explore those limits within a multi-objective framework that seeks to optimize any measure of accuracy and fairness and provides a Pareto front with the best feasible solutions. In this work, we focus our study on decision tree classifiers since they are widely accepted in machine learning, are easy to interpret and can deal with non-numerical information naturally.

We conclude experimentally that our method can optimize decision tree models by being fairer with a small cost of the classification accuracy. We believe that our contribution will help stakeholders of socio-technical systems to assess how far they can go being fair and accurate, thus serving in the support of enhanced decision making where machine learning is used.
\end{abstract}


\end{frontmatter}


\section{Introduction}

Algorithmic and data-driven decision making has rapidly swept through several social, political and industry contexts. Beyond the possible misuses of technology, there is an increased awareness that these processes are not neutral and can reproduce and amplify past and current structural inequalities \cite{oneil_weapons_2016,eubanks_automating_2018}. Within this context, particular interest is paid to the role of machine learning (ML) with well known examples of models biased against historically discriminated groups \cite{supreme2009ricci, angwin2016machine, bolukbasi_man_2016} or the intersection of these groups \cite{kearns_preventing_2017,buolamwini_gender_2018}. Fairness in ML has emerged as a community initially motivated to develop technological solutions to the disparate impact and treatment by biased algorithms~\cite{zafar2017fairness,XIA2019,zehlike_matching_2020, lipton_does_2017, bolukbasi_man_2016} that also moves to a broader and multi-disciplinary understanding of the issues of socio-technological interventions \cite{binns_its_2018,mitchell_model_2019,selbst_fairness_2019,sanchez-monedero_what_2020}. This work contribute to this field by studying how far bias mitigation can go whilst satisfying the accuracy and transparency of the models, thus providing a tool for a wider understanding of the technological boundaries of socio-technical proposals.




Bias mitigation techniques can broadly be divided into three non-exclusive categories \cite{friedler2019comparative}: (1) preprocessing, (2) inprocessing, and (3) postprocessing. The preprocessing techniques attempt to learn new representations of data to satisfy fairness definitions. The inprocessing methods involve modifying the classifier algorithm by adding a fairness constraint to the optimization problem. The postprocessing methods aim at removing discriminatory decisions after the model is trained. Normally, in inprocessing approaches the fairness criteria are used as an optimization constraint rather than as a guide to build a more equitable prediction model. As a result of the optimization process, those fixed restrictions will come out with a degree of equity that might not match the problem requirements whereas the space of solutions that can be reached remains unknown so that decision makers cannot observe the range of possibilities and their behaviour.  



The main contribution of this paper is a methodology that explores optimal ML solutions and evaluates the boundaries of fairness in relation to other dimensions of the evaluation of an ML model. We claim that multi-objective evolutionary algorithms might be used to direct a meta-learning process for optimizing the hyperparameters of a classifier. In particular, we focus the study on the suitability of decision trees as base learners because of their properties of transparency and accuracy. Thus, we propose to use a genetic algorithm to tune the decision tree hyperparameters to find models that offer a wide repertoire of balances between precision and fairness. The architecture of this methodology can be applied on any type of classifier and hyperparameter set and the optimization is independent of the definition of fairness and precision. As a result of the meta-learning process, the method produces a Pareto front with a set of optimal feasible solutions. In this way, the method addresses the before mentioned issues of single constrained optimization proposals to build fair models.

We conduct an extensive set of experiments based on 5 real-world datasets which are widely used in the fairness literature. The solution space obtained by our approach indicates that there exists a wide number of optimal solutions (Pareto optimal) that are characterized by not being dominated by each other. We also evaluate the boundaries between accuracy and fairness that can be achieved on each problem, giving an empirical visualization of the limits between both measures. In addition, we assess how decision trees hyperparameters are affected by this tradeoff. Finally, a convergence analysis is also presented in order to evaluate the evolutionary approach of this methodology.

As far as we know, multi-objective optimization has not yet been used in the field of fairness in ML, so we believe that the proposal will open a very fruitful and beneficial research line, enriching the state-of-the-art. 

\section{Background}
To ground our methodology, we begin by reviewing relevant related works. We then introduce evolutionary algorithms and briefly explain a type of algorithm belonging to this family.

\subsection{Optimizing for fairness and accuracy}

Bias mitigation algorithms often explicitly or implicitly add fairness constraints on model group performance. In this section, we introduce some related works that aim at optimizing for fairness and accuracy. For further information on the relation between accuracy and fairness measures we refer to \cite{menon_cost_2018}.

In the context of decision trees, in \cite{kamiran_discrimination_2010} the information gain function is modified for splitting and pruning to add the entropy with respect to the sensitive attribute. The authors explored several options. The first one considers the entropy with respect to the class label, but it does not allow splitting if it introduces discrimination with respect to sensitive attribute. 
The second alternative implements a tradeoff between objectives by dividing the gain in accuracy by the gain in discrimination. This option did not achieve suitable results.

More recently, authors in \cite{agarwal_reductions_2018} proposed to reduce fair classification to a sequence of cost-sensitive classification tasks to obtain Pareto optimality between overall accuracy and any fairness definition. In a related work, authors in \cite{balashankar_pareto-efficient_2019} find a Pareto optimal point which maximizes multiple subgroup accuracy measures while satisfying equality of opportunity.

Zafar et. al \cite{zafar_fairness_2019} formulated the problem as a convex constrained optimization problem that allows a dual formulation in which accuracy is optimized under fairness constraints. In their formulation, fairness is introduced in terms of a measure of the decision boundary fairness that serves as a proxy to many fairness statistical metrics. The tradeoff between accuracy and fairness due to disparate mistreatment is expressed as a threshold parameter established by the user. Moreover, the formulation allows introducing several attributes as constraints, e.g. race and gender.

Hu et al.~\cite{hu_fair_2019} transformed the constrained loss minimization problem into a social welfare maximization problem. Using SVM's regularization path and techniques from parametric programming, they show that always preferring more fair solutions does not abide by the Pareto Principle. They concluded that applying strict fairness criteria can lead to worse welfare outcomes for the groups. 

Thus, there is an interest in exploring the simultaneous optimization of accuracy and fairness metrics. Some proposals obtain Pareto optimal solutions that implicitly set a tradeoff between objectives, whereas others relly this on a user parameter. As an alternative to this, our work aims to provide the whole Pareto front as a means to explore the impact of the ML models, or, in general, to understand the behaviour of the combination between a dataset and knowledge representation.



\subsection{Evolutionary algorithms}
Multi-objective optimization is a field of decision making which aims at optimizing simultaneously more than one objective function. This field of research has developed a large number of applications in engineering, economics, and logistics where optimal decisions need to be taken in the presence of tradeoffs between two or more competitive objectives. Maximizing comfort and energy saving in a climatization system is a practical example of multi-objective  problem involving two objectives. Mathematically, this can be formulated as:

\begin{equation*}
\text{min } (f_1(x), \ldots, f_n(x)) \text{   } \text{ s.t. } x \in X,
\end{equation*}

\noindent where $n > 1$ is the number of objective functions and $X$ is the set of feasible solutions.

When multiple objective functions appear in a problem, no single solution exists that optimizes each function at once. Otherwise, the presence of multiple objectives gives a set of optimal solutions, possibly infinite. A solution is \textit{non-dominated} whether does not exist another solution that dominates the current one, i.e., it does not improve one objective function without worsening other objective functions. Formally:

\begin{defn}
A solution $x \in X$ is said to \textit{dominate} another solution $x' \in X$, if it is better or equal in all the objectives and strictly better in at least on of them, i.e.:
\begin{itemize}
    \item $f_i(x) \preceq f_i(x')\text{, } \forall i \in \{1, \ldots, n\} $ and,
    \item $f_j(x) \prec f_j(x')\text{, } \text{for at least one index } j \in \{1, \ldots, n\}$.
\end{itemize}
\label{def:non-dominated}
\end{defn}
A solution is called \textit{Pareto optimal} if there does not exist another solution that dominates it. Consequently, the set of all  Pareto optimal solutions is defined as \textit{Pareto front} or \textit{boundary}. Assessing this frontier allows decision makers to select any efficient solution, depending on the worthiness of each objective function.

Evolutionary algorithms (EAs) are often well-suited for solving optimization problems. They consist of meta-heuristic-based methods inspired by some aspects of natural evolution. The basic idea is that unfit members will die and not contribute to the gene pool of the offspring, while fitter individuals are allowed to survive and contribute to generate new solutions. Over the last decades, a number of multi-objective EAs have been developed, capable of searching for multiple Pareto optimal solutions concurrently in a single run.


\subsection{NSGA-II}
The non-dominated Sorting Genetic Algorithm (NSGA)~\cite{srinivasan1994multi} was one of the first EAs developed for multi-objective problem optimization. Yet this approach was criticized due to: (1) the high computational complexity, (2) the lack of elitism, and (3) the low spread of solutions. Then, the NSGA-II~\cite{deb2002fast} was proposed as a modification to address these disadvantages. To solve (1) the authors proposed a \textit{non-dominated} sorting procedure where all the individuals are sorted according to the level of non-dominance. In order to address the issue (2), they implemented \textit{elitism} to store all non-dominated solutions and help to prevent the loss of good solutions once they are found. This aspect also enhances the convergence property of EAs~\cite{zitzler2000comparison}. Finally, they adapted a suitable automatic mechanism based on the \textit{crowding distance} to ensure diversity in a population and then solve (3). This distance function assigns a distance metric to all individuals within a population and then compares whether two solutions are close enough. A solution with a smaller value is more crowded by other solutions, therefore is more likely to not survive in further populations.


This approach starts by creating an initial parent population $P$ of size $N$ randomly. The population is evaluated by the objective functions and sorted following the non-dominance criteria~\ref{def:non-dominated}. After that, a rank score is assigned to each solution where the first level corresponds to the best individuals, the second level is the next-best set of members, and so on.  After that, the binary tournament selection, crossover, and mutation operators are used to create an offspring population. These children are also evaluated by the objective functions and combined together with the previous population. All individuals are then ranked and sorted by the non-domination rank and the crowding distance, which is considered the elitist step. The $N$-best members are then selected to form the next population.  Finally, the algorithm finalizes when last generation is reached. 

\section{Multi-objective method for accurate and fair machine learning}

Our proposed methodology is based on a multi-objective algorithm with a generational evolutionary approach. The goal is to guide a ML classifier to obtain the best tradeoffs between accuracy and fairness (Pareto optimal), by learning the best combination of hyperparameters given a dataset and its protected attribute that identifies a protected group. The selection mechanisms are inspired by the elitist NSGA-II method~\cite{deb2002fast} which was described in the previous section. In particular, we propose decision trees as the ML classifiers to be optimized due to its comprehensive and transparency nature.

\subsection{Meta-learning approach}

\begin{figure*}
\centering
    \includegraphics[trim=7 4 6 3,clip,width=.85\textwidth]{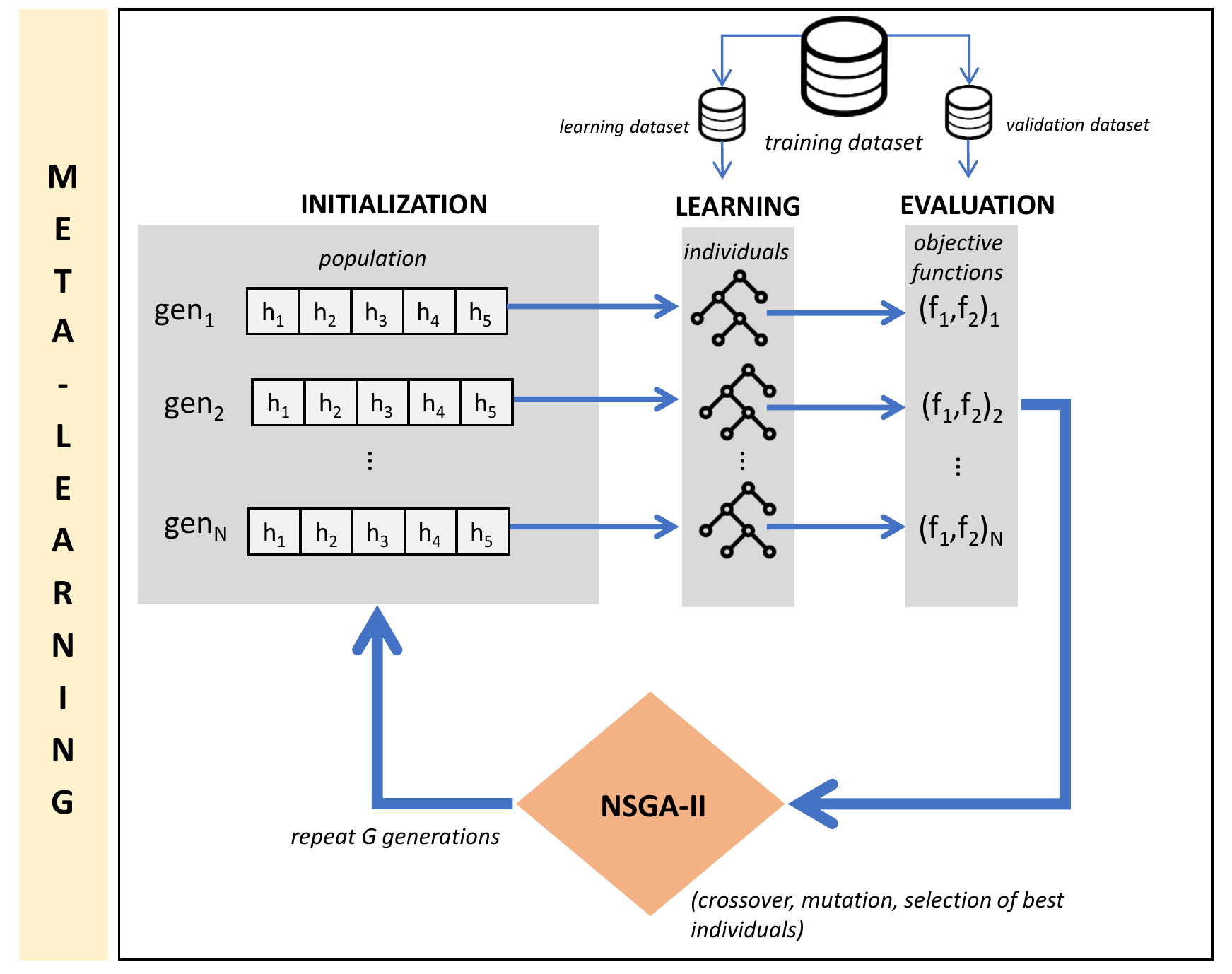}
    \caption{This diagram overviews the flow of the proposed meta-learning. The first population is randomly generated at the inicialization step. Given the values of each gene, $N$ decision trees are trained and evaluated with each combination of hyperparameters afterwards. The NSGA-II ranks the individuals, i.e. the trained decision trees, by evaluating the objective functions on the validation set. After that, the NSGA-II generates an offspring population which is also evaluated. Finally, the method selects the best $N$-members among parents and children to form the next population using a selection mechanism known as \textit{elitist non-dominated sorting}. This process is repeated until the last generation $G$ is reached.}
    \label{fig:nsga-diagram}
\end{figure*}

The pseudo-code of the meta-learning approach is presented in Algorithm~\ref{alg:meta-learning}. Additionally, Figure~\ref{fig:nsga-diagram} presents a visual diagram of the process.

Specifically, the meta-learning consists of
dividing the training set into two subsets (learning and validation) where the decision tree will be built from the first set, and the measures will be evaluated in the second one. The multi-objective algorithm will ensure that, iteration after iteration, the set of the best hyperparameter configurations will survive so that the NSGA-II will explore other settings around them. At the end, a set of optimal solutions is returned which will be tested in the testing set.

In this work, we propose decision trees as the base classifiers for the meta-learning. Decision trees are considered \textit{white box} models, since it is easy to analyze the steps taken to classify data~\cite{rudin2018please}. They are easy to interpret, and they can be summarized in a set of rules. In addition, these kind of algorithms do not require data normalization or dummy variables creation, since they are able to use both numerical and categorical data. This fact simplifies the preprocessing step, which can directly affect the accuracy and fairness of the classifier~\cite{friedler2019comparative}. Particularly, from these ML algorithms we are interested in the following hyperparameters:

\begin{itemize}
\item \texttt{criterion}: This function measures the quality of a split. Decision trees split nodes as long as this value decreases. The purity of a node can be measured with the \textit{Gini} index and the \textit{entropy}.
\item \texttt{max\_depth}: The maximum depth of the tree. Deeper trees are more complex.
\item \texttt{min\_samples\_split}: The minimum number of samples required to split an internal node. In that case, a higher number of samples implies simpler trees.
\item \texttt{max\_leaf\_nodes}: Total number of leaves in a tree. The higher the number of leaves, the more complex the tree.
\item \texttt{class\_weight}: It is used to give weight to each class, which is considered when measuring the quality of the splits. It is very useful for unbalanced datasets where models usually misclassified the minority class. It takes values in $[0,1]$. The positive class is weighted with \texttt{class\_weight}, while the negative one is $1-$\texttt{class\_weight}. A value of $0.5$ means both classes are evenly considered.
\end{itemize}

The \texttt{criterion}, \texttt{max\_depth}, and \texttt{min\_samples\_split} adjust the size of the tree in different directions, which means that different balances between precision and complexity can be found. Moreover, if the search of the best set of hyperparameters is guided by any fairness metric, the structure of the tree can be regulated towards branches that do not generate disparities among groups. The \texttt{class\_weight} hyperparameter addresses disparity by the transferring instances between false positives and false negatives.

The main advantage of the proposed method is that it can obtain a wide number of optimal solutions in just one run. The design of this methodology also allows the use of any ML classifier without the need to modify it, unlike the inprocessing techniques. Moreover, our methodology supports any type of data since decision tree classifiers allow either numerical and categorical attributes. Finally, our proposal also supports another of the fairness community's claims, which is transparency. By using decision trees as classifiers, we allow decision makers to know all the decisions the model made once trained. Also, optimization functions do not need to be differentiable allowing a wider bank of fairness definitions.

Formally, the $j$th-individual, $I_{kj}$, of the $k$th-population, $P_k$, is
a trained decision tree. In turn, this tree is trained with a $m$-tuple \textit{gen}, $g_{kj}$, which contains the values of
each hyperparameter $h = \{h_1, \ldots, h_m\}$ on each corresponding position, hence $m = 5$:

\noindent $$I_{kj} := \text{\texttt{decision\_tree}}(g_{kj})$$
\begin{equation*}
    \begin{aligned}
h := \{ \text{\texttt{criterion, max\_depth, min\_samples\_split,}} \text{\texttt{ max\_leaf\_nodes, }} \\ \text{\texttt{class\_weight}} \}.
    \end{aligned}
\end{equation*}
Since some of the tree hyperparameters are categorical or integer numbers, the chromosomes are decoded after its generation in order to obtain the proper value for the classifier.

\begin{algorithm}
\caption{Meta-learning algorithm}
\begin{algorithmic}
\REQUIRE objective function of accuracy and fairness ($f_1$ and $f_2$), number of hyperparameters of the ML classifier ($m$), intervals of hyperparameters ($\text{min } (h_i)$ and $\text{max } (h_i) \text{ } \forall i \in \{1, \ldots, m\}$), datasets, and the protected attribute
\ENSURE Set of ML models with different accuracy-fairness tradeoffs
\DATA training (learning and validation) and testing dataset ($D_{learn}$, $D_{val}$, and $D_{test}$)
\PARAMETERS number of generations ($G$), population size ($N$), crossover probability ($p_c$), mutation probability ($p_{m}$), mutation parameter ($\mu$)
\BEGIN
\STATE initialize population $P_{1}$
\STATE evaluate objective functions ($P_{1}$, $D_{val}$)
\STATE non-dominated rank individuals of $P_{1}$
\WHILE{$k \leq G$}
\STATE $P^{(1)}_{k} \leftarrow$ elitist selection ($P_{k-1}$)
\STATE $P^{(2)}_{k}\leftarrow$ crossover ($P^{(1)}_{k}$)
\STATE $P^{(3)}_{k}\leftarrow$  mutation ($P^{(2)}_{k}$)
\WHILE{$1 \leq l \leq N$}
\STATE create $S_{kl}$ solution by training classifier ($I_{kl}$, $D_{learn}$)
\STATE evaluate objective functions ($S_{kl}$, $D_{val}$)
\ENDWHILE
\STATE non-dominated rank individuals of population $P^{(3)}_{k}$
\STATE $P_{k} \leftarrow$ elitist non-dominated replacement ($P^{(3)}_{k}$, $P_{k-1}$)
\ENDWHILE
\STATE \textbf{return} non-dominated solutions in $P_k$
\END
\end{algorithmic}
\label{alg:meta-learning}
\end{algorithm}

In the following sections we extensively describe the evolutionary algorithm components.

\subsection{Pool initalization}
The initialization step generates the first pool. The first individual generated ($I_{11}$) is created with default values of hyperparameters: $g_{11} = (Gini, \infty, 2, \infty, 0.5)$. The purity of the node is measured with the \textit{Gini} index; the tree can be widened and deepened as needed since the limits for the depth and number of leaves within a node is not fixed and the lowest minimum of samples to split is used; both positive and negative class have the same weight. After training the first tree with these hyperparameters, the remaining individuals are generated considering the actual values of depth and leaves of that first tree as limit. The second individual will be generated with entropy criterion and those limits, while the rest of individuals are generated with random hyperparameters within the limits fixed by the first individual.

For a better understanding of the previous paragraph, we propose a practical case. Given the first individual of the first generation of the meta-learning ($I_{11}$), the first tree is trained with the specific values of the hyperparameters $(Gini, \infty, 2, \infty, 0.5)$. Thereafter, the decision tree has a depth of value $depth(I_{11}) = D$ and a total number of leaves equals to $leaves(I_{11}) = L$. The second individual ($I_{12}$) is then trained with the following hyperparameter set: $(entropy, D, 2, L, 0.5)$. These limits for the depth and number of leaves of the tree ($D$ and $L$) will be preserved throughout the process until completion, i.e., $I_{1j} = (c,d,s,l,w)$ with $c \sim \{Gini,entropy\}$, $d \sim U(1,D)$, $s \sim U(2,\texttt{training\_set\_size})$, $l \sim U(1,L)$, and $w \sim U(0,1)$. In this way, this adhoc modification will let the meta-learning to better adjust to dataset characteristics.

\subsection{Crossover operator}

The crossover generates two individuals ($I_{kj}$ and $I_{k,j+1}$) that inherit the hyperparameters given by two parents ($I_{k-1,a}$ and $I_{k-1,b}$), depending on the \textit{crossover probability} ($p_c$). Concretely, this match is based on a given parameter $u \sim \mathcal{U}(0,1)$ which follows a uniform distribution. If this value is $u \leq p_c$, the crossover function assigns the same hyperparameter value of the parents to the children. Otherwise, it assigns a linear combination of parents' hyperparameters ($g_{k-1,a}$ and $g_{k-1,b}$), where the parameter $\beta \sim \mathcal{U}(0,1)$:

$$g_{kj} = \frac{g_{k-1,a} + g_{k-1,b}}{2} + \beta \frac{|g_{k-1,a} - g_{k-1,b}|}{2}$$
$$g_{k,j+1} = \frac{g_{k-1,a} + g_{k-1,b}}{2} - \beta \frac{|g_{k-1,a} - g_{k-1,b}|}{2}$$

After that, genes of the resulting offspring are rounded off and decoded in order to obtain the proper values for the hyperparameters. In integer genes, the rounded values replaces the decimal ones to ensure a more effective search space.

\subsection{Mutation operator}

The mutation operator changes the real membership function hyperparameter values encoded in the
chromosome, according to the \textit{mutation probability} ($p_m$) per individual. The gene (hyperparameter) to be mutated is randomly selected over the five genes. Then, given $u', u'' \sim \mathcal{U}(0,1)$, the
chromosome is mutated as follows:

\begin{equation*}
\begin{aligned}
g_{kj}=\left\{
             \begin{array}{ll}
             g_{kj} + \delta (g_{kj}-\text{min }(h_i)), & u' < 0.5 \\
             g_{kj} + \delta (\text{max }(h_i)- g_{kj}), & u' \geq 0.5
             \end{array}
\right.
\end{aligned}
\end{equation*}

\noindent where,

\begin{equation*}
\begin{aligned}
\delta=\left\{
             \begin{array}{ll}
             -1+2u''^{\frac{1}{\mu+1}}, & u'' \leq 0.5 \\
             1-2(1-u'')^\frac{1}{\mu+1}, & u'' > 0.5.
             \end{array}
\right.
\end{aligned}
\end{equation*}


\subsection{Multi-objective approach}

The multi-objective optimization is based on two objective functions to be minimized: $f_1$ evaluates the accuracy and $f_2$ the fairness of the model. Thus, $f_1$ is focused on improving the prediction performance while $f_2$ is used to mitigate the discrimination of the ML algorithm.

Both concepts of accuracy and fairness can be defined in several ways, referring to different meanings. Although the proposed methodology is totally flexible for using any definition, in this work we focus on two of them. We define $y$ as the binary class label vector where 1 is the positive outcome and 0 is the negative outcome; $\hat{y}$ is the predicted outcome of the ML classifier; $z$ is the associated protected feature of each individual, where 1 is the privileged class.

\subsubsection{Error}
We consider the \textit{Geometric Mean} (G-mean) to evaluate the performance of the assessment task. G-mean is also widely used for quantifying the classifier performance in class imbalanced problems, since it evaluates both positive and negative class. It combines True Positive Rate (TPR) ($Pr(\hat{y} = 1 \mid  y = 1 )$) and True Negative Rate (TNR) ($Pr(\hat{y} = 0 \mid  y = 0 )$):
$$
\text{G-mean}(\hat{y}, y) = \sqrt{P(\hat{y} = 1 \mid  y = 1 ) \cdot P(\hat{y} = 0 \mid  y = 0 )}.
$$

By maximizing this measure, we ensure the cost of false positive and false negative to be low. Since our method is designed for a minimization problem, we consider the first objective function as the G-mean error, i.e. $f_1(\hat{y}, y) = 1-\text{G-mean}(\hat{y}, y)$.

\subsubsection{Unfairness}
We consider the difference of the unfairness measure proposed for avoiding \textit{disparate mistreatment}, defined as \textit{False Positive Rate} (FPR)~\cite{zafar2017fairness, chouldechova2017fair}. This definition ensures that missclassification rates are balanced across groups of the protected attribute $z$:
$$
f_2(\hat{y}, y) = \text{FPR}_{\text{diff}}(\hat{y}, y) = \lvert P(\hat{y} \neq y \mid z = 0, y = 0) - P(\hat{y} \neq y \mid z = 1, y = 0)  \lvert .
$$

\subsubsection{Domination criterion}

Given $X$ the genotype (hyperparameters) and $Y$ the phenotype (decision trees), the $f:X\rightarrow Y$ map obtained by the proposed method is characterized by being a non-injective non-surjective function.  It is not injective as different values of hyperparameters can lead to obtain exactly the same decision tree. It is not surjective as the image (set of all possible decision trees generated by our method) does not fill the whole codomain, i.e., it is not possible to obtain any decision tree, only those generated by the learner. The cardinality of $Y$ is much more lesser than the cardinality of $X$.

As a result, there are many different individuals that generate exactly the same decision tree, and so the same objective functions. This impairs the search process as variations generated by crossover and mutation do not change the objective functions. To palliate this effect, we have improved the domination criterion as follows. Once two individuals have the same values for both objectives, we consider that the individual that generates the tree with the lowest number of leaves dominates the other one. In case of a tie also in this value, the individual with the lowest value of the hyperparameter \texttt{max\_leaf\_nodes} is considered to dominate the other one.



\section{Experimental Analysis}

In this section we first describe the datasets used for assessing the proposed methodology. After that, we define the parameter setup used in these experiments. Finally, the obtained results and its analysis is provided.

\subsection{Datasets}

We ran experiments based on five realworld datasets from different domains like salaries, recruitment processes, credit risks, or recidivism risk assessment. These datasets have been widely used as benchmarking in state-of-art in fairness~\cite{friedler2019comparative}. They are freely available in a Github repository\footnote{\url{https://github.com/algofairness/fairness-comparison/tree/master/fairness/data}. Last date accessed: June 9, 2020}. A brief description of the dataset context is given below:

\begin{itemize}
    \item[-] \textbf{Adults:} This dataset contains demographic information about US citizens in 1994\footnote{\url{http://archive.ics.uci.edu/ml/datasets/adult}}. There are 32,561 instances and 14 attributes. The prediction task is to asses whether an individual earns more (positive class) or less (negative class) than \$50K per year. The protected attribute considered is \textit{race}.
    \item[-] \textbf{German:} It contains financial information about individuals\footnote{\url{http://archive.ics.uci.edu/ml/datasets/statlog+(german+credit+data)}}. There are 1,000 instances and 20 attributes. The prediction task is to assess the credit risk of individuals. The protected attribute considered is \textit{age}.
    \item[-] \textbf{ProPublica:} This dataset is about the performance of COMPAS algorithm, a statistical method for assigning risk scores within the US criminal justice system created by Northpointe. It was published by ProPublica in 2016~\cite{angwin2016machine}, claiming that this risk tool was biased against African-American individuals. In this dataset, they analyzed the COMPAS scores for ``risk of recidivism'' and checked to see how many were charged with new crimes over the next two years. It contains individuals from the Broward County (Florida) in 2013 and 2014. There are 7,214 individuals containing 52 attributes. From these attributes, we have used the following 12 in the experiments of this paper~\cite{friedler2019comparative}: \texttt{sex, age, age\_cat, race, juv\_fel\_count, juv\_misd\_count, juv\_other\_count, priors\_count, c\_charge\_degree, c\_charge\_desc, decile\_score, score\_text}. The prediction variable is whether the individual will be rearrested in two years or not. The protected attribute is \textit{race}.
    \item[-] \textbf{ProPublica violent:} This dataset describes the same scenario as the previous one, but in this case the outcome is whether the rearrest happened within two years was for a violent crime~\cite{angwin2016machine}. It contains 4,743 individuals and also the 12 attributes. The protected attribute is also \textit{race}.
    \item[-] \textbf{Ricci:} This dataset comes from labour law case from the United States, where several firefighters from New Haven (Connectitut, US) claimed for disparate impact on the  promotion decision. It contains the scores obtained in the  exam  taken to be promoted~\cite{supreme2009ricci}. There are a total number of 118 rows and 4 attributes. The protected attribute is \textit{race}.
\end{itemize}

Each dataset is preprocessed to assure that the input data satisfies the classifier requirements by removing features that should not be used for the classification task, imputing missing values or transforming features like dates, etc. We also transform all the protected attributes into binary (e.g., ``white''-``not white'', ``younger than 25 years old''-``older than 25 years old'', ``caucasian''-``not caucasian'').
Table~\ref{tab:summary_datasets} shows the number of features selected for each dataset and class distribution.

\begin{table}
  \caption{Summary of datasets.}
  \label{tab:summary_datasets}
  \begin{tabular}{rccc}\hline
    \textbf{Dataset} & \# \textbf{Features} & \textbf{\# Positive} & \textbf{\# Negative} \\ \hline
    \textbf{Adults} & 14 & 7,841 & 24,720 \\
    \textbf{German} & 20 & 300 & 700  \\
    \textbf{ProPublica} & 12 & 3,251 & 3,963 \\
    \textbf{ProPublica Violent} & 12 & 775 & 3,968 \\
    \textbf{Ricci} & 4 & 56 & 62  \\ \hline
\end{tabular}
\end{table}


\subsection{Parameter setup}

The experiments are replicated 10 times with different seeds to ensure stability and reproducibility. In each seed, the learning (56.25\%), validation (18.75\%), and testing (25\%) sets are randomly sampled. The parameters for the evolutionary method are set as follows:
\begin{itemize}
    \item 300 generations ($G = 300$),
    \item 50 individuals ($N = 50$),
    \item 1 as crossover probability ($p_c = 1$),
    \item 0.3 as mutation probability ($p_{m} = 0.3$),
    \item 5 as mutation parameter ($\mu = 5$).
\end{itemize}

The code is implemented in Python using libraries such as \texttt{pandas} for data processing, \texttt{sklearn.DecisionTreeClassifier} for machine learner (CART algorithm) and \texttt{numpy} for numerical processing. The original code of the NSGA-II algorithm is available at \url{github.com/baopng/NSGA-II} (last date accessed: June 9, 2020). This research complies with research reproducibility principles. Code and data are made open and available in a public repository: \url{https://github.com/anavaldi/fairness_nsga} (last date accessed: June 9, 2020).

\subsection{Analysis of results}

In this section, we empirically study the limits of the accuracy-fairness tradeoff. We first analyze the properties of the Pareto optimal solutions obtained when optimizing both together. We also analyze the relationship between decision tree learner's hyperparameters and measures' values. Finally, we present the convergence properties of the meta-learning approach.

\subsubsection{Analysis of accuracy-fairness tradeoff}
\label{sec:fairness_boundaries}

\newcommand*\rot{\rotatebox{90}}

\setlength{\tabcolsep}{1.8mm}
\renewcommand{\arraystretch}{0.95}

\begin{table}
\caption{\label{tab:summary_Pareto_frontier}Averaged distribution of error ($1-$G-mean) and unfairness (FPR$_{\text{diff}}$) measures in the obtained Pareto optimal solutions for validation ($v$) and test ($t$) datasets. Depth and leaves (complexity of the models) are the actual values of the generated decision trees. This table reflects the accuracy-fairness tradeoff (i.e., \emph{how fair can we go}) in each real-world problem}
\begin{tabular}{crcccccc}
\hline
 &  & Error$_v$ & Unfairness$_v$ & Error$_t$ & Unfairness$_t$ & Depth & Leaves\\
\hline
\multirow{5}{*}[-0.5ex]{\rot{\textbf{Adult}}} & \textit{min} & .17644 & .06743 & .18238 & .07218 & \hphantom{0}8.4 & \hphantom{0}95.5\\
 & \textit{Q$_1$ (25\%)} & .19374 & .04036 & .19412 & .05822 & 12.1 & 211.2\\
 & \textit{Q$_2$ (50\%)} & .21715 & .02423 & .22220 & .04577 & 14.5 & 352.4\\
 & \textit{Q$_3$ (75\%)} & .26488 & .00971 & .26804 & .02620 & 16.8 & 518.9\\
 & \textit{max} & .35766 & .00034 & .35759 & .00794 & 22.6 & 945.6\\
\hline
\multirow{5}{*}[-0.5ex]{\rot{\textbf{German}}} & \textit{min} & .26780 & .12406 & .32393 & .16990 & \hphantom{0}6.9 & \hphantom{0}22.3\\
 & \textit{Q$_1$ (25\%)} & .27830 & .08135 & .34387 & .13916 & \hphantom{0}7.9 & \hphantom{0}28.1\\
 & \textit{Q$_2$ (50\%)} & .29442 & .04411 & .35488 & .11279 & \hphantom{0}9.1 & \hphantom{0}34.0\\
 & \textit{Q$_3$ (75\%)} & .31977 & .01989 & .37343 & .07821 & \hphantom{0}9.4 & \hphantom{0}40.2\\
 & \textit{max} & .38101 & .00099 & .43214 & .02597 & 10.3 & \hphantom{0}47.6\\
\hline
\multirow{6}{*}[-0.5ex]{\rot{\textbf{ProPublica}}} & \textit{min} & .32759 & .12471 & .33676 & .12871 & \hphantom{0}6.7 & \hphantom{0}50.5\\
 & \textit{Q$_1$ (25\%)} & .34078 & .08052 & .35094 & .08936 & 10.0 & 145.2\\
 & \textit{Q$_2$ (50\%)} & .35572 & .03476 & .36223 & .07011 & 12.1 & 238.4\\
 & \textit{Q$_3$ (75\%)} & .38492 & .01362 & .39121 & .04591 & 14.4 & 312.0\\
 & \textit{max} & .39997 & .00293 & .40881 & .03026 & 16.7 & 467.4\\
 & COMPAS & .35002 & .12519 & .34759 & .14751 & --- & ---\\
\hline
\multirow{6}{*}[-0.5ex]{\rot{\textbf{ProPublica Violent}}} & \textit{min} & .31366 & .10367 & .33176 & .10261 & \hphantom{0}6.2 & \hphantom{0}34.6\\
 & \textit{Q$_1$ (25\%)} & .33651 & .06047 & .35422 & .07879 & \hphantom{0}8.9 & \hphantom{0}71.7\\
 & \textit{Q$_2$ (50\%)} & .35388 & .03446 & .37430 & .05461 & 10.8 & 109.6\\
 & \textit{Q$_3$ (75\%)} & .38638 & .01011 & .41021 & .03251 & 12.2 & 148.2\\
 & \textit{max} & .48942 & .00021 & .50264 & .01794 & 14.4 & 210.1\\
 & COMPAS & .32388 & .13474 & .33494 & .13897 & --- & ---\\
\hline
\multirow{5}{*}[-0.5ex]{\rot{\textbf{Ricci}}} & \textit{min} & .04487 & 1.0000 & .12249 & .80222 & \hphantom{0}1.8 & \hphantom{0}\hphantom{0}2.9\\
 & \textit{Q$_1$ (25\%)} & .09006 & .71526 & .15782 & .66326 & \hphantom{0}2.1 & \hphantom{0}\hphantom{0}3.4\\
 & \textit{Q$_2$ (50\%)} & .13134 & .46007 & .18936 & .54931 & \hphantom{0}2.4 & \hphantom{0}\hphantom{0}3.8\\
 & \textit{Q$_3$ (75\%)} & .17195 & .30838 & .21820 & .43881 & \hphantom{0}2.6 & \hphantom{0}\hphantom{0}4.1\\
 & \textit{max} & .21268 & .15669 & .24781 & .32831 & \hphantom{0}2.9 & \hphantom{0}\hphantom{0}4.4\\
\hline
\end{tabular}
\end{table}

The averaged results over 10 runs are shown in Table~\ref{tab:summary_Pareto_frontier} for the five real-world problems. To represent the average distribution of the obtained results, we have computed the average of the ten runs at minimum value of error in validation dataset (Error$_v$), 25th percentile (Q$_1$), 50th (Q$_2$), 75th (Q$_3$) and maximum value of error. As the set of inferred solutions are Pareto efficient, the corresponding values of unfairness are reversely sorted. In the case of the two ProPublica problems, the results obtained by COMPAS are also included to better understand the room for improvement in those cases.

The obtained results in Ricci are very particular. We found that this problem is very easy to be solved in terms of accuracy by a decision tree learner, i.e., it is possible to obtain solutions with almost zero error and, therefore, almost one unfairness. In fact, in some partitions the solution found was perfect. Consequently, the multi-objective optimization tends to obtain very spread Pareto solutions, so we decided to leave this problem out of the rest of the analysis.

While the validation dataset is used to guide the meta-learning algorithm, the test dataset is never used. When comparing validation and test columns, we observe that, although the scores in test are slightly worse than validation (as expected), the Pareto efficiency in test also remains, which shows the robustness of our methodology. Yet the results are overfitted regarding the unfairness measure (i.e., strong differences between Unfairness$_v$ and Unfairness$_t$ in Table~\ref{tab:summary_Pareto_frontier}) when it comes to very low values.

When comparing the average results that occupy the first (\textit{min}) and 50th (Q$_2$) positions of error (Q$_1$ in Adult), we are able to estimate the percentage of accuracy that needs to be sacrificed to improve fairness. The accuracy lost in test (Error$_t$) is 6\%, 10\%, 8\% and 13\% in Adult, German, ProPublica, and ProPublica Violent problems, respectively, whilst the fairness improvement is of 81\%, 66\%, 54\% and 53\%, respectively. This gives us an idea of how it is possible to optimize the ML process to generate fairer solutions without an excessive loss of precision, which should encourage ML designers to incorporate fairness criteria into these processes.

Focusing on the two ProPublica problems, where the prediction made by COMPAS is widely known, we can analyze the accuracy and fairness achieved by the Northpointe's software when assessing a criminal defendant's likelihood to re-offend. We can observe that the most unfair solution got by our methodology is much fairer than the obtained by COMPAS. This demonstrates the improvement margin of fairness in these problems when guiding the ML process by unbiased measurements. If we interpolate the fairness scores got by our methodology for an accuracy equal to COMPAS's, the test results would be $(\text{Error$_t$}, \text{Unfairness$_t$}) = (0.3476,0.0987)$ in ProPublica and $(\text{Error$_t$}, \text{Unfairness$_t$}) = (0.3349,0.0992)$ in ProPublica Violent, showing that our method improves the fairness of COMPAS's solutions in 67\% and 71\%, respectively, without compromising accuracy.

When analyzing the performance of solutions, we are additionally concerned with transparency of the classifiers. Indeed, in the problems considered in our experimental analysis, where wrong outcomes may discriminate unfavored social groups, to understand the reasoning behind a machine decision is critical. Therefore, we analyze in  which degree the Pareto optimal models are also easy to interpret. The fact of being using a decision tree structure to represent the knowledge helps to understand the machine decision criteria compared with other black-box models, but the complexity of these trees will also influence on its interpretability, as an excessively fine-grain decision boundary (high number of leaves) and complex multivariate conditions (high depth of the tree) would be hardly understandable.

Analyzing the complexity results in Table~\ref{tab:summary_Pareto_frontier}, we observe that the number of leaves is relatively low in the most accurate solutions, but tends to increase as fairness improves. This effect shows that the method needs to use more leaves to improve fairness with a minimum loss of accuracy. This is an expected result since equalizing false positive rates between the two people groups forces a finer partitioning of data. The high depth with a relatively low number of leaves suggests the construction of unbalanced decision trees (keep in mind than a perfectly balanced binary tree would need $2^\text{depth}$ leaves, which is very far from what we get). That is, some few leaves need a high depth (i.e., extensive multiple conditions) to be effective.

Analogously, it is well known that a lower error implies a higher complexity, so it is curious to observe that this relation is not shown in the obtained results. The reason is simply that the complexity (number of leaves and depth) is not considered as a criteria to be optimized by our methodology, so this variable is freely adapted to the two contradictory objectives (accuracy and fairness), both of them demanding higher complexity to be reached. It seems that the fairness objective ends up winning the battle. In other words, the algorithm finds it harder to improve fairness than accuracy with a reduced complexity. Nevertheless, this interesting effect deserves a deeper study that would divert us from the main goal of our research in this paper, so we leave it as a further research line.

\begin{figure*}[t!]
    \centering
    \begin{subfigure}{0.50\textwidth}
        \includegraphics[trim=24 10 44 35,clip,width=\textwidth]{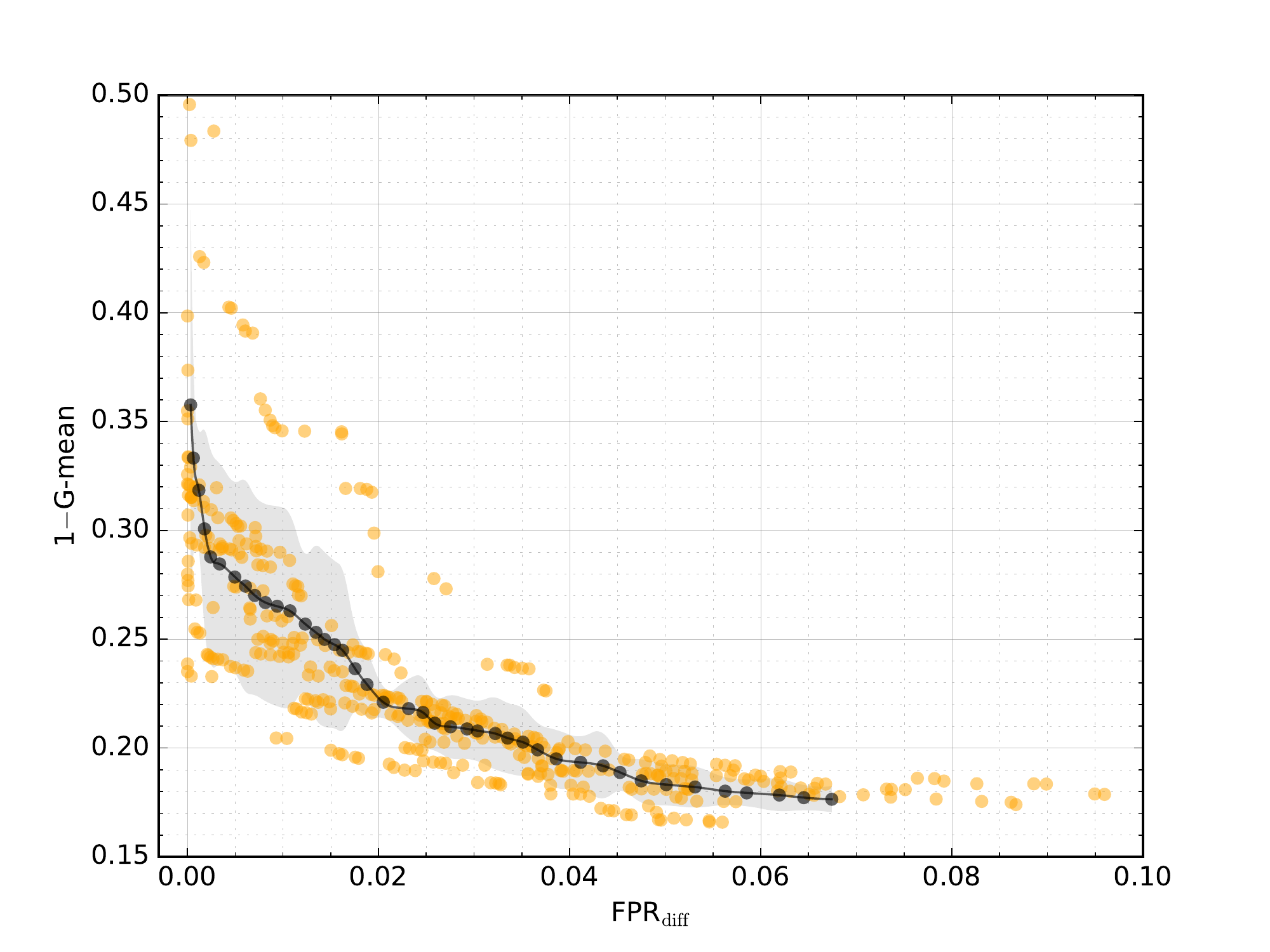}\vspace{-2mm}
        \subcaption{Adult}
        \label{fig:sub_a_distance_by_authors}
    \end{subfigure}%
    \begin{subfigure}{0.50\textwidth}
        \includegraphics[trim=24 10 44 35,clip,width=\textwidth]{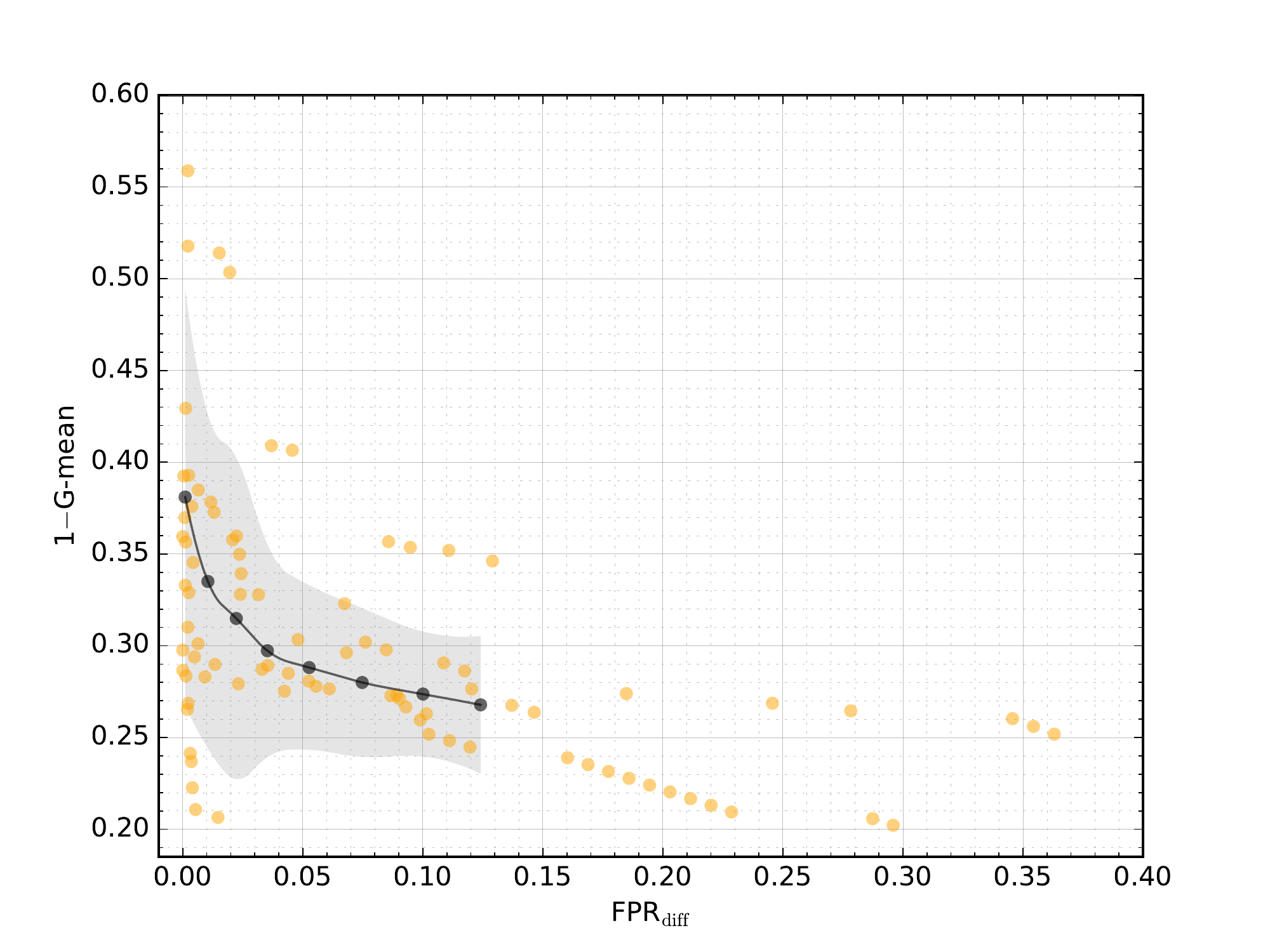}\vspace{-2mm}
        \subcaption{German}
        \label{fig:sub_b_distance_by_authors}
    \end{subfigure}\\[2mm]
    \begin{subfigure}{0.50\textwidth}
        \includegraphics[trim=24 10 44 35,clip,width=\textwidth]{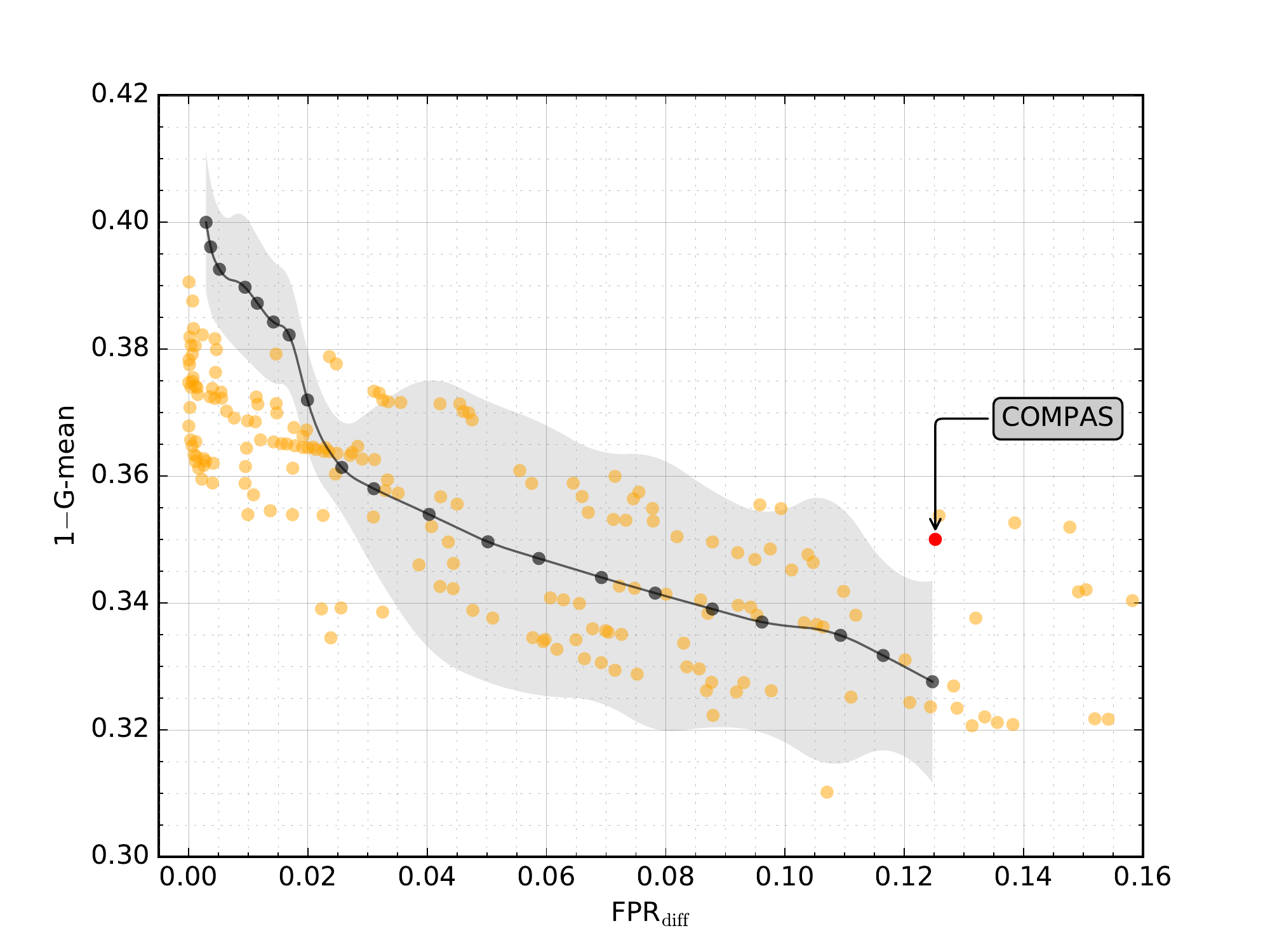}\vspace{-2mm}
        \subcaption{ProPublica}
        \label{fig:sub_c_distance_by_authors}
    \end{subfigure}%
    \begin{subfigure}{0.50\textwidth}
        \includegraphics[trim=24 10 44 35,clip,width=\textwidth]{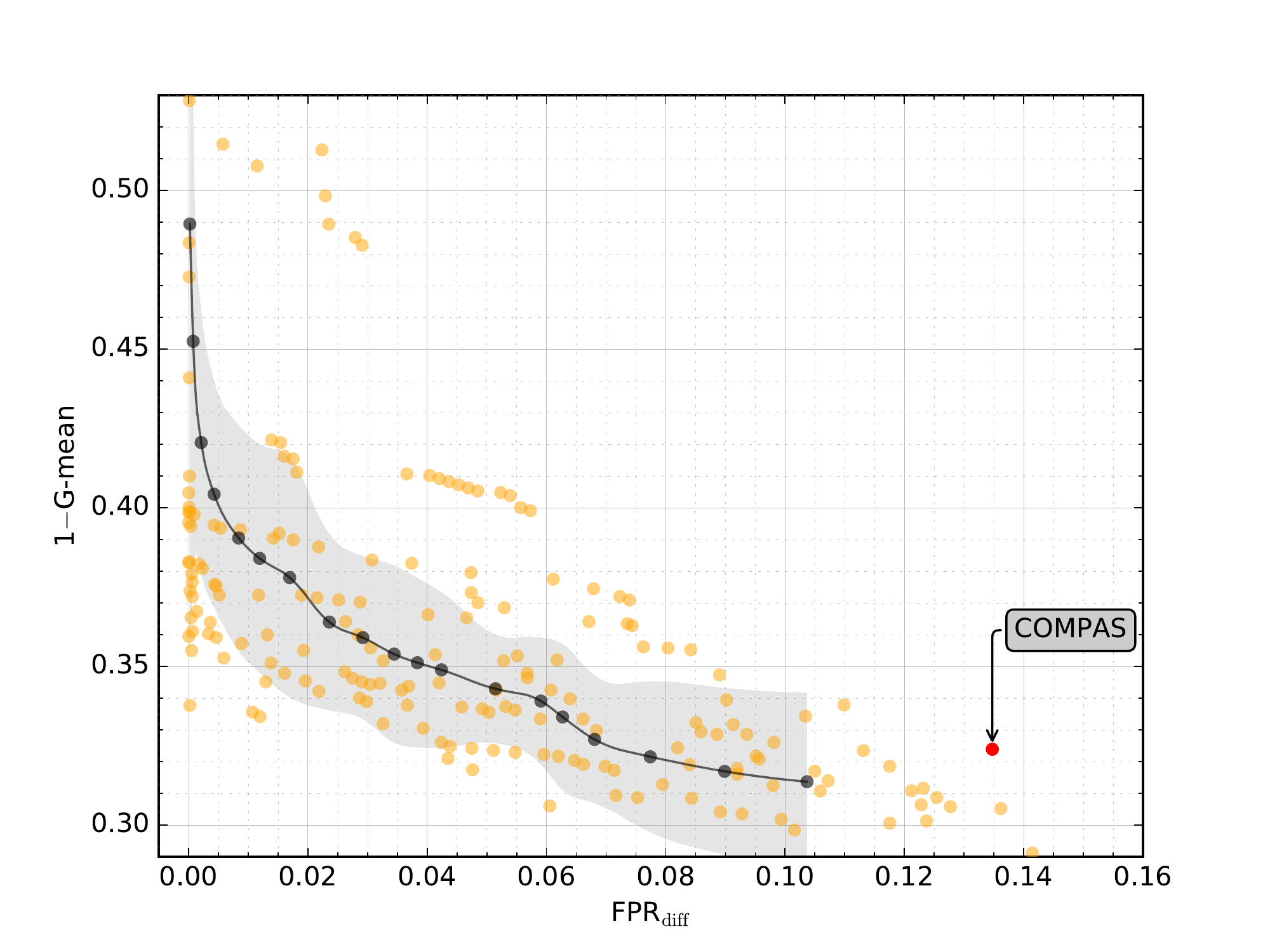}\vspace{-2mm}
        \subcaption{ProPublica Violent}
        \label{fig:sub_d_distance_by_authors}
    \end{subfigure}
    \caption{Orange dots represent Pareto optimal solutions---minimizing error ($1-$G-mean) vs. unfairness (FPR$_\text{diff}$)---found by the proposed algorithm in different problems. Dark gray dots indicate the average Pareto set, which is a way of representing \textit{how fair can we go} in a specific problem or, in other words, which shape takes the accuracy-fairness tradeoff. Light gray area is the interquartile range. Our methodology is effective to find a wide spread of solutions that are accurate and fair at the same time. In the two ProPublica datasets, the meta-learning algorithm also finds better solutions than the obtained by COMPAS (red dots), showing that there is a wide range of possibilities to be fairer without worsening accuracy}
    \label{fig:non-dominated solutions}
\end{figure*}

To better understand the behavior of the proposed method, Figure~\ref{fig:non-dominated solutions} plots the obtained Pareto optimal solutions, with orange dots being the solutions of each run and dark gray dots connected by lines represent the average Pareto front. This average Pareto is obtained by firstly getting the rounded mean number of different solutions $n$ (which corresponds to the number of dark gray dots) and then obtaining the average values at $n$ different percentiles positions equally distributed. For example, if we have three runs where we got 3, 5 and 7 Pareto optimal solutions, we would obtain the $n=5$ evenly distributed percentiles (i.e., 1st, 25th, 50th, 75th and 100th) with linear interpolation between adjacent ranks in each run and then calculate the average value for each percentile. The interquartile range (Q$_3$$-$Q$_1$) of the error is represented with the light blue area in the figures.

The spread of the dots (especially in fairness dimension) and the width of the interquartile range suggests us that the attainable levels of accuracy and fairness is quite sensitive to the dataset partitions into training and test. This is particularly serious in German. The exception is represented by Adult, where the solutions in different data partitions are very compact. This may be due the fact that Adult has a considerably high number of data, so that the bias of the data partitioning is mitigated. As our methodology splits the training data into learning and validation, it suffers when very little data is available, as in German.

As we can observe from the plotted Pareto fronts, the contradictory condition between accuracy and fairness is clear: more accuracy implies less fairness, and vice versa, as analyzed in previous works \cite{agarwal_reductions_2018,balashankar_pareto-efficient_2019}. Although what is really interesting to analyze is the shapes of the averaged Pareto fronts as they provide valuable information about how the combination dataset and decision tree is working. In fact, beyond generating a wide repertoire of solutions with different balances of accuracy and fairness, our methodology also returns a greater understanding of the problem by explaining how these contradictory criteria are related.

Let us take as example the ProPublica problem. The accuracy-fairness relation is rather linear in the range $[0.026,0.125]$ of unfairness, i.e., range $[0.328,0.361]$ of error. Then, we see a clear knee of the curve below an unfairness of 0.026, meaning beyond this threshold, improving a bit the fairness has a relatively high cost in accuracy. Similar conclusion can be taken in the other problems, where the unfairness threshold is around 0.01 in Adult, and 0.02 in German and ProPublica Violent. This knowledge could be used by other researchers and practitioners to set different fairness requirements depending on the problem when employing decision trees.

\subsubsection{Analysis of learner's hyperparameters}

\begin{figure*}[t!]
    \centering
    \begin{subfigure}[b]{0.495\textwidth}
        \includegraphics[trim=24 29 10 37,clip,width=\textwidth]{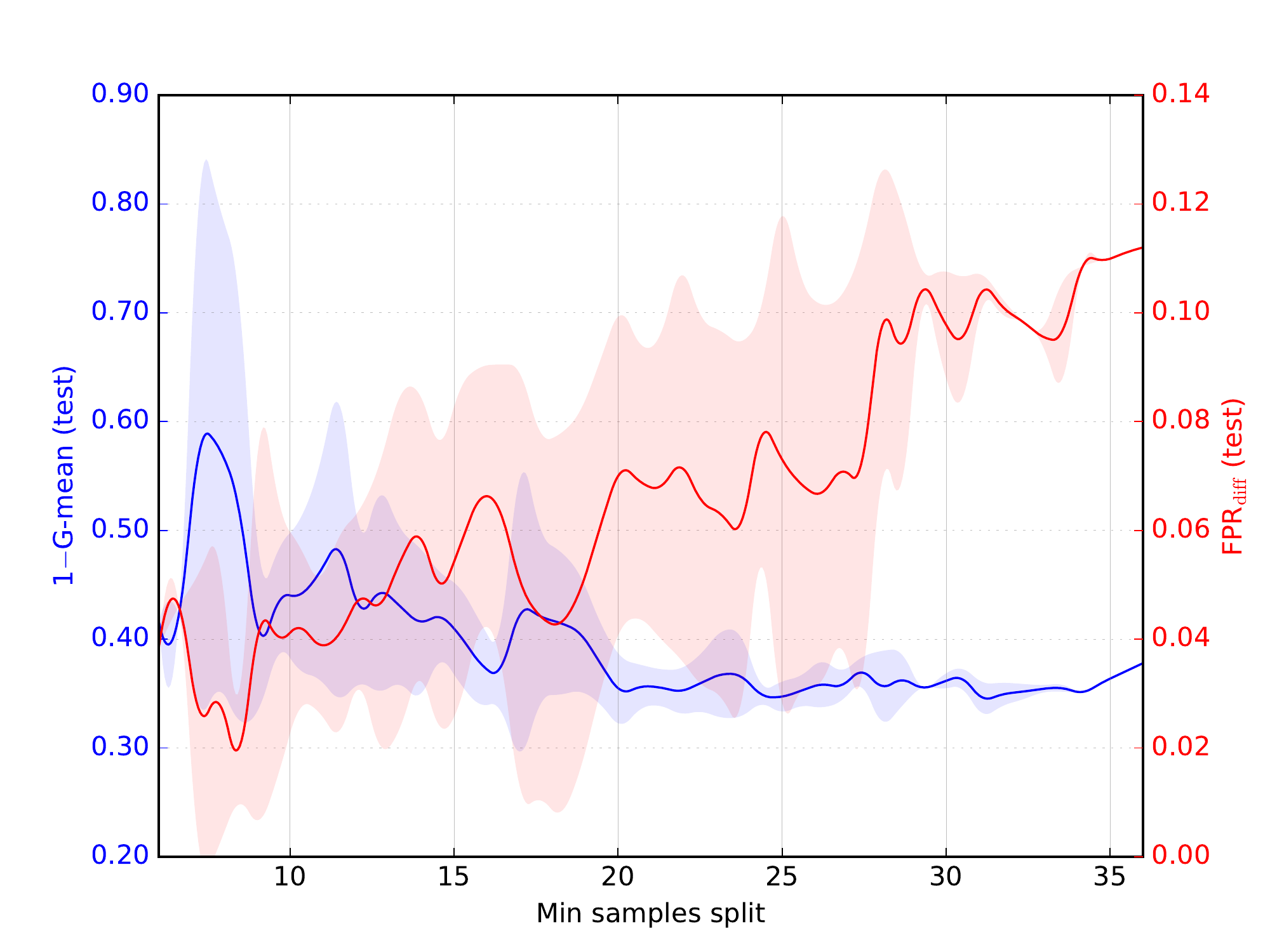}
        \subcaption{Min samples split}
        \label{fig:sub_maxdepth}
    \end{subfigure}\hfill
    \begin{subfigure}[b]{0.495\textwidth}
        \includegraphics[trim=24 29 10 37,clip,width=\textwidth]{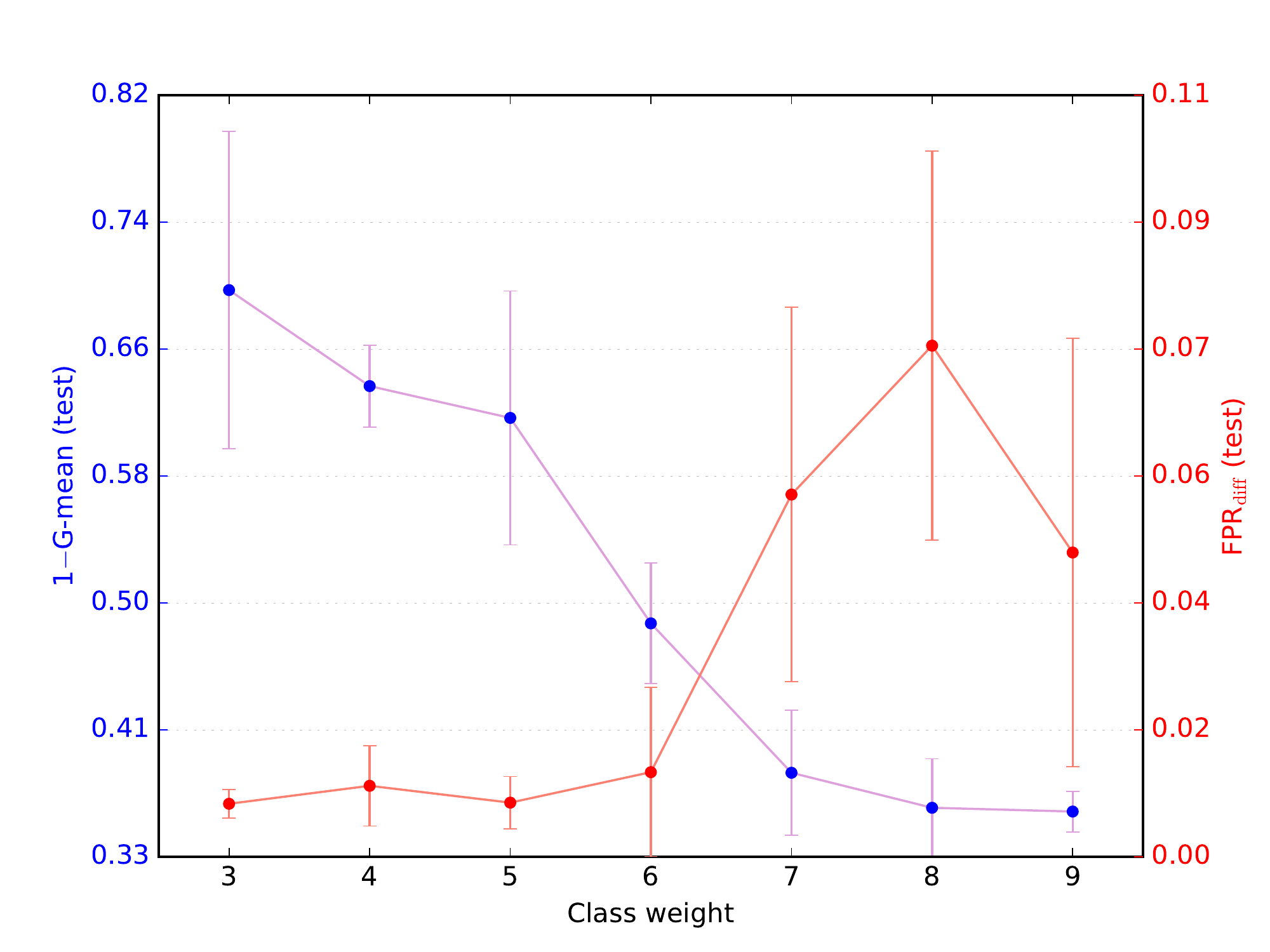}
        \subcaption{Class weight} 
        \label{fig:sub_criterion}
    \end{subfigure}
    \caption{Effect of two hyperparameters of the decision tree learner in ProPublica Violent. Notice how different values of them impacts varyingly on accuracy and fairness. Low threshold to split a node and low weight of the positive (minority) class favor the generation of decision trees with a good fairness}
    \label{fig:hyperparameters}
\end{figure*}

As we are proposing a meta-learning method that indirectly controls the generated decision tree by tuning the hyperparameters of the learner, we are also interested in assessing the impact of learner's hyperparameters on the performance. We have already discussed in the previous section the effect of demanding optimal fairness in the complexity of the trees (good fairness needs higher number of leaves). Here we analyze the effect of two other hyperparameters: \texttt{min\_samples\_split} and \texttt{class\_weight}. We did not find significant results in the fifth hyperparameter (\texttt{criterion}). Figure~\ref{fig:hyperparameters} shows the values of these two hyperparameters in the obtained Pareto optimal solutions of ProPublica Violent. The mean values over all the runs is plotted as lines and dots, while the shaded areas and error bars correspond to the standard deviation. Blue color is used for error and red color for unfairness, both in test datasets (Error$_t$ and Unfairness$_t$).

In the case of \texttt{min\_samples\_split}, the results confirm our guess that in order to improve fairness it is necessary to deepen certain branches of the tree, so that a low value of the limit of samples needed to divide a node helps to generate fairer trees. It is interesting to see here how a high value of this limit hurts fairness a lot but does not influence accuracy.

With regard to \texttt{class\_weight}, which controls the importance of the positive class (and reversely the negative one), the effect is as follows. In accuracy, a higher weight of the positive class implies generating more accurate solutions in this imbalance dataset (there are five times more of the negative than the positive). This makes sense as G-mean measure rewards balanced predictive precision in the two classes, so making more important the minority (positive) class helps to this goal. This hyperparameter has the contrary effect in fairness. Here, fairer solutions are obtained when a positive class weight in $[3,5]$ is given (moreover, with a low variance that ensure statistical significance), i.e., to decrease the importance of the positive class (which in the analyzed problem means that the criminal defendant re-offends) reduces the false positives, which makes easier to generate decision trees with a better balance of false positive rates between the two groups (Caucasian vs. rest of ethnics). In other words, giving less credibility to the positive class (re-offend) allows for fairer classifiers. However, we cannot ignore that this could also be a side effect of the Pareto efficiency followed by the optimization process.

\subsubsection{Analysis of convergence}

An algorithm converges when there is no significant improvement in the values of the objective functions of the population from one to the next generation. This aspect is important to be studied in order to assess efficiency of the method. At the same time, its analysis can reveal the resistance of each problem to allow improvements of the accuracy and fairness measures.

\begin{figure*}[t!]
    \begin{subfigure}[b]{0.5\textwidth}
        \includegraphics[trim=24 29 46 37,clip,width=\textwidth]{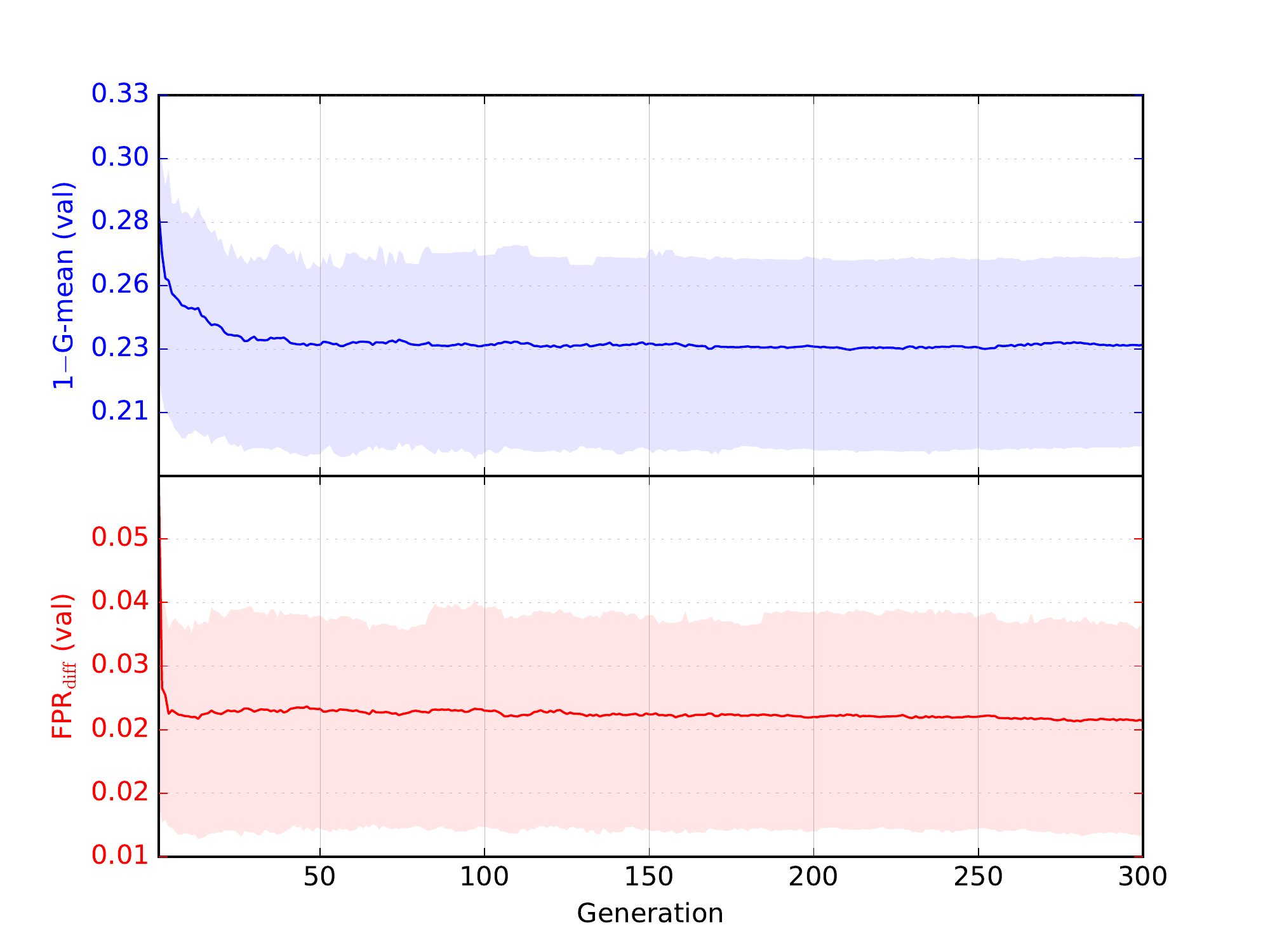}\vspace{-2mm}
        \subcaption{Adult}
        \label{fig:conv_adult}
    \end{subfigure}\hfill
    \begin{subfigure}[b]{0.5\textwidth}
        \includegraphics[trim=24 29 46 37,clip,width=\textwidth]{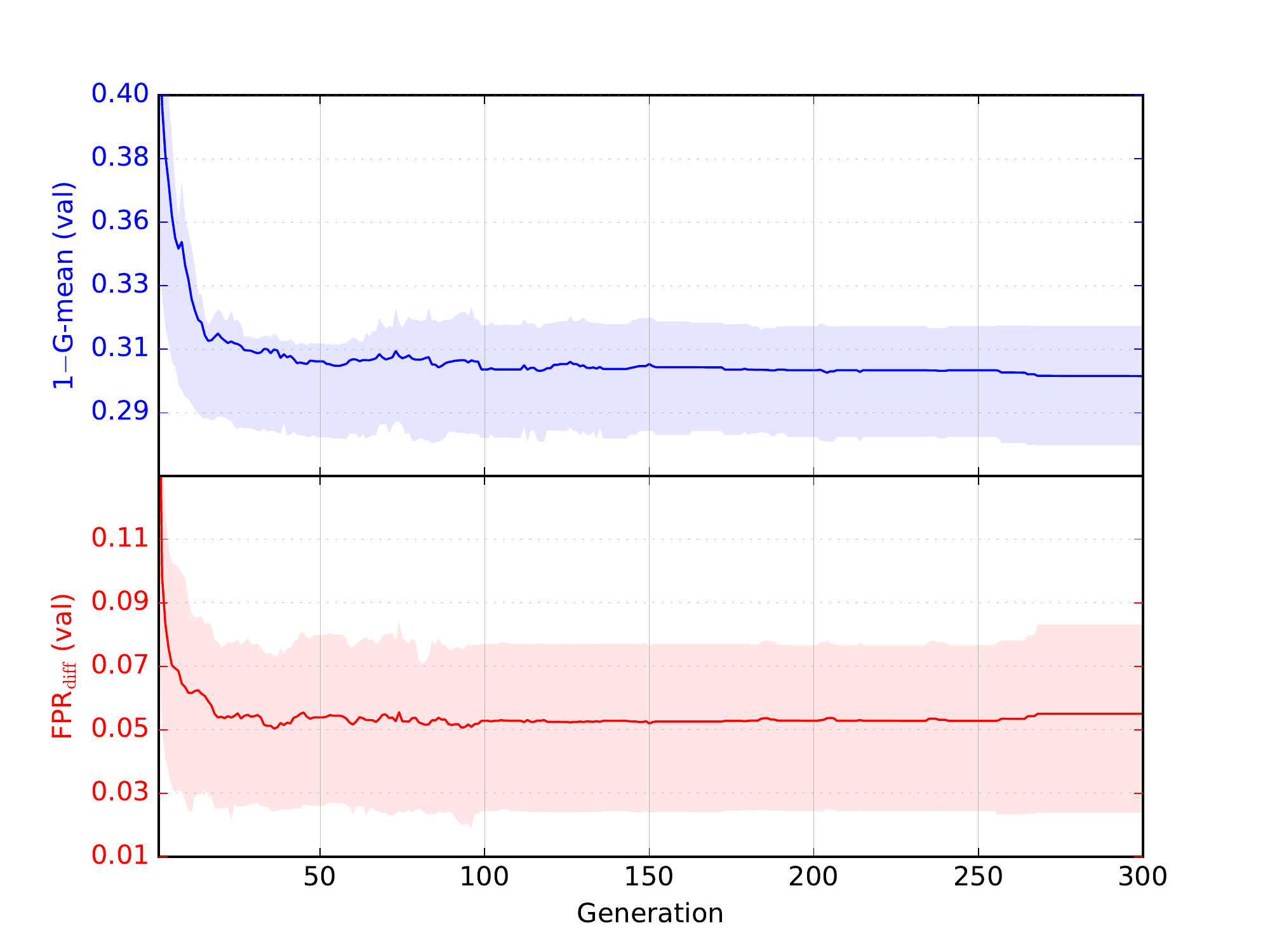}\vspace{-2mm}
        \subcaption{German}
        \label{fig:conv_german}
    \end{subfigure}\\[2mm]
    \begin{subfigure}[b]{0.5\textwidth}
        \includegraphics[trim=24 29 46 37,clip,width=\textwidth]{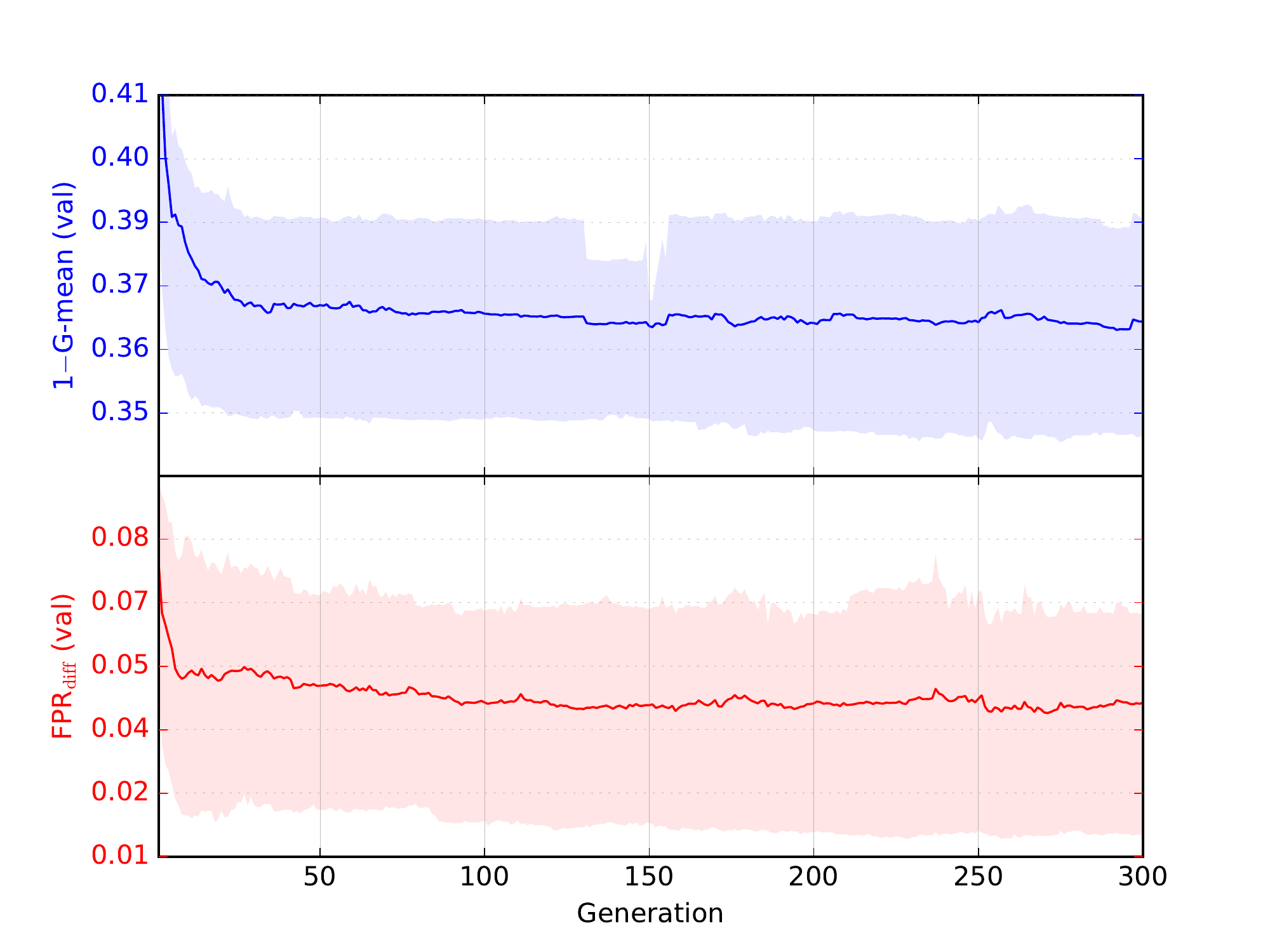}\vspace{-2mm}
        \subcaption{ProPublica}
        \label{fig:conv_pro}
    \end{subfigure}\hfill
    \begin{subfigure}[b]{0.5\textwidth}
        \includegraphics[trim=24 29 47 37,clip,width=\textwidth]{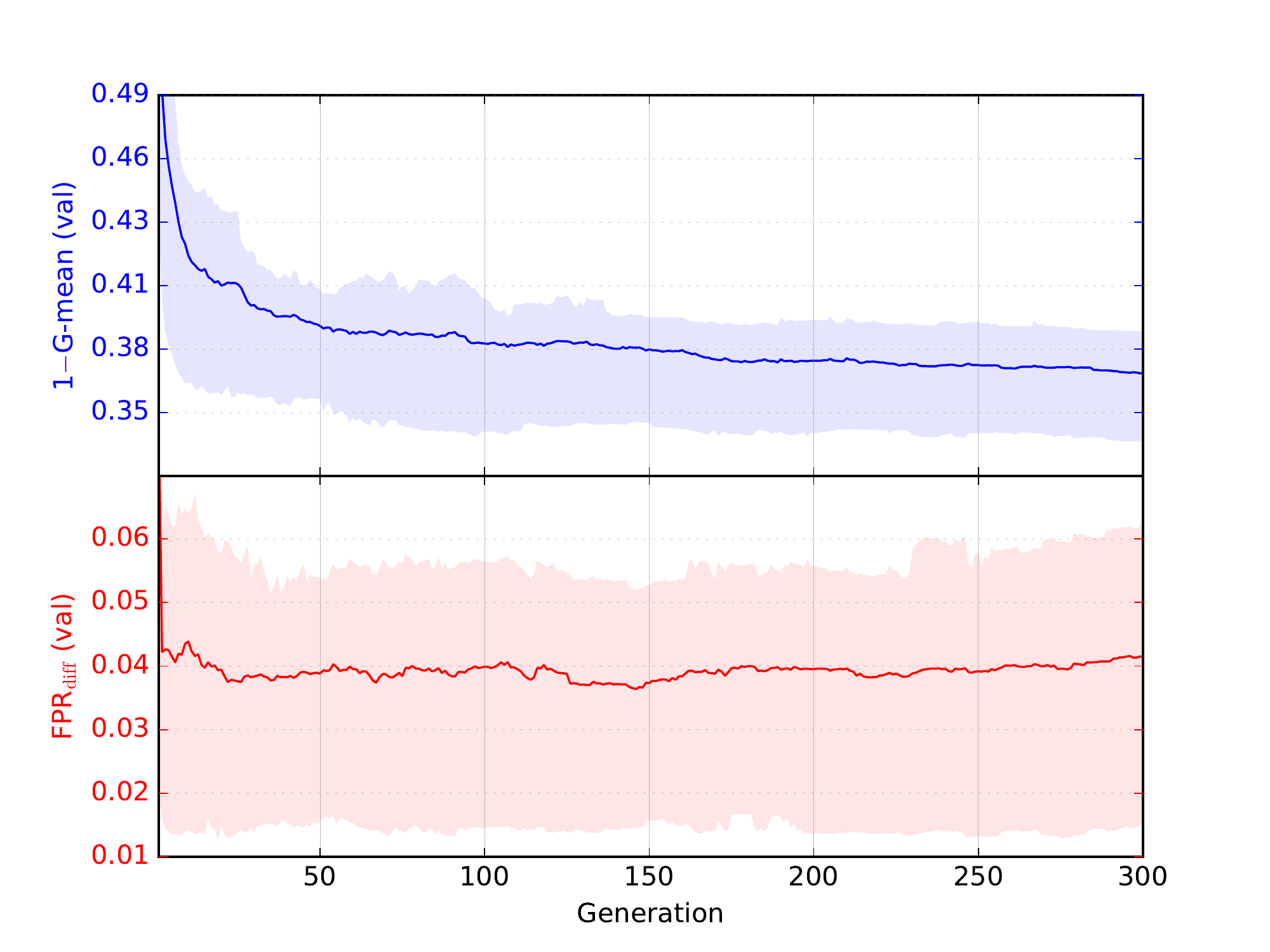}\vspace{-2mm}
        \subcaption{ProPublica Violent}
        \label{fig:conv_pro_violent}
    \end{subfigure}
    \centering
    \caption{Evolution of non-dominated solutions through 300 generations of the meta-learning algorithm. To represent the distribution of these Pareto sets, mean (line) and Q$_1$-Q$_3$ (area) of error and unfairness objectives (i.e., in validation set) averaged over 10 runs are plotted.
    }
    \label{fig:convergence}
\end{figure*}

In multi-objective optimization, convergence is more complex to analyze as many optimal solutions evolve at the same time. To summarize the behavior of the process, Figure~\ref{fig:convergence} presents the mean, Q$_1$ and Q$_3$ of the two objectives (error and unfairness in validation set) for the obtained Pareto set at each generation (averaged results over 10 runs are plotted). In some way, the mean gives an idea about the quality of the solutions (the lower the better) while the interval $[\text{Q}_1,\text{Q}_3]$ represents the diversity of the Pareto sets (the wider the better). In Ricci, the algorithm fully converges very quickly (these values do not change at all after 27 generations), so we omit this plot for the sake of clarity of the paper.

We observe that low unfairness is faster to get than low error, so in the first third of the evolution good fairness is reached in all the problems, while the accuracy is slowly improved until the end of the process. Adult has the most stable convergence of the four shown problems due to the reduced bias in data partitions as above said. German also converges very well, but with a slight improvement of accuracy in the last 40 generations at the expense of making fairness slightly worse. ProPublica shows the most continuous convergence where accuracy and fairness are persistently improved. In ProPublica Violent, good fairness is very quickly obtained while accuracy is continually enhanced.

\section{Conclusions}

In this work we propose a meta-learning multi-objective optimization algorithm to explore  the boundaries of fairness in real world problems. We present a methodology that (1) enables standard ML algorithms to be fairness-aware, (2) obtains the experimental frontier of the accuracy-fairness tradeoff, (3) uses interpretable models as base learners to comply with transparency values, and (4) converges rapidly to optimal solutions. To the best of our knowledge, this is the first work that proposes both accuracy and fairness as objective functions for a multi-objective ML approach.

\textit{Accuracy-Fairness}: Throughout the experimental analysis, we show the optimal fitness that can be achieved by optimizing the geometric mean of the predictive precision of each class versus false positive rate equality of the groups, i.e., no disparate mistreatment as defined in \cite{zafar_fairness_2019}. The cost in accuracy when satisfying fairness criteria has been theoretically studied (e.g. \cite{menon_cost_2018,zafar_fairness_2019}). These studies demonstrate the existence of a unavoidable tradeoff between accuracy with respect to the target variable and fairness with respect to the sensitive attribute. That is, when one objective is improved by the model the second one is penalized. Or what is the same, these two objectives are contradictory. Based on this assertion, we design in this paper an optimization process able to push both objectives to the frontier where the mentioned Pareto efficiency is reached, thus returning a plethora of solutions with different accuracy-fairness balances. Besides, the experimental analysis shows how fair can we go in a specific problem by decision trees, providing further insight about the capability of standard ML algorithms to get good fairness and the flexibility of the problem (dataset) to allow this.



\textit{Fairness-Transparency}: As it is well known in ML, decision trees can improve accuracy (at least, while the sweet spot without overfitting is reached) often by increasing the model complexity (i.e., tree depth and number of leaves). Moreover, we believe that, in order to improve fairness, the decision tree needs to be deeper for a fine-grain data partition to hold misclassification parity between different groups having different values of the sensitive attribute. Therefore, both accuracy and fairness demand more complex decision trees. When optimizing accuracy and fairness together, we find that the process tends to solve the conflict by generating more complex trees in fairer solutions even when their accuracies are not so good. This may be due to the fact that, when optimizing learner's hyperparameters as our methodology do, fairness is mainly reached by more complex trees while there are other chances of improving accuracy by fine tuning the remaining hyperparameters.

\textit{Convergence}: Evolutionary algorithms are sharply criticized because of its low convergence in many problems. Nevertheless, we show that this methodology early achieves optimal solutions. Objective functions cooperate to generate good solutions in the first generations, but they compete to obtain optimal solutions at the end.



\textit{Future work}: Although we know that technology interventions alone will not address social injustice, there are several interesting directions highlighted by our findings. From the obtained results, it is clear a further research is needed to understand the role of transparency (in terms of model complexity) in the accuracy-fairness tradeoff. Therefore, we propose to add the complexity of the trees as a third objective function ($f_3$). Regarding the fact that fairness can be defined in multiple ways, we plan to develop further analysis with different measures of mistreatment. In relation to claims by \cite{menon_cost_2018}, it would be interesting to study dataset properties, such as correlation of the sensitive attribute with the target variable. 
We are aware that the experiments presented in this work only include one binary sensitive attribute. We propose to consider more attributes in further experiments to analyze how convergence is affected. Differential fairness~\cite{foulds2020intersectional} is a growing concept highly related with this work, which addresses intersectionality. We propose to run new experiments of our meta-learning algorithm proposing this new fairness definition. Finally, it is worth mentioning that our approach is completely flexible, and its design allows the use of any type of classifier and hyperparameters, that serving as a tool to experimentally analyze several dimensions of the behavior of ML methods.



\bibliography{mybibfile}

\end{document}